\documentclass{article} % For LaTeX2e
\usepackage{iclr2025_conference,times}

\usepackage{wrapfig}
\usepackage{lipsum} % 用于生成示例文字
\usepackage{wrapfig}

\usepackage{amsmath,amsfonts,bm}
\usepackage{multicol}
\def\eqref#1{equation~\ref{#1}}
\def\1{\bm{1}}

\DeclareMathAlphabet{\mathsfit}{\encodingdefault}{\sfdefault}{m}{sl}
\SetMathAlphabet{\mathsfit}{bold}{\encodingdefault}{\sfdefault}{bx}{n}

\usepackage{booktabs}      % 提供更美观的表格线 (如 \toprule, \midrule, \bottomrule)
\usepackage{multirow}      % 用于创建跨多行的单元格
\usepackage{amssymb}       % 提供了丰富的数学符号，包括 \checkmark (对勾)
\usepackage{graphicx, subfigure}     % 用于处理图片和缩放内容 (\resizebox)

\usepackage{colortbl}
\usepackage{xcolor}

\usepackage{hyperref}
\usepackage{url}
\usepackage{multicol}
\usepackage{aliascnt}
\usepackage{adjustbox}
\usepackage{tcolorbox}
\usepackage{siunitx} % for S column type in tables
\usepackage{booktabs}
\usepackage{multirow} 
\usepackage{enumitem}
\usepackage{booktabs} % 用于更美观的表格线
\usepackage{amssymb}  % 用于 \checkmark
\usepackage{amsmath}  % 用于 \texttimes (或者使用 pifont 和 \ding{55})
\usepackage{graphicx} % 可能需要，如果调整列宽
\usepackage{longtable} % 如果表格太长需要跨页
\usepackage{graphicx}
\usepackage{subcaption}
\usepackage{calc}

\usepackage{CJKutf8} % 支持中文
\definecolor{uclablue}{rgb}{0.15, 0.45, 0.68}
\hypersetup{
    breaklinks,
    citecolor=uclablue,
    colorlinks=true,
}

\newtcolorbox{AIbox}[2][]{aibox,title=#2,#1}
\tcbset{
  aibox/.style={
    top=10pt,
    colframe=black,
    colbacktitle=black,
    center
  }
}

\title{Adaptive Termination for Multi-round Parallel Reasoning: An Universal \\ Semantic Entropy-Guided Framework}

\author{
Zenan Xu$^1$, Zexuan Qiu$^{1,2}$, Guanhua Huang$^1$, Kun Li$^{1,2}$, Siheng Li$^{1,2}$, Chenchen Zhang$^1$ \\
Kejiao Li$^1$, Qi Yi$^1$, Yuhao Jiang$^1$, Bo Zhou$^1$\thanks{Correspondence to Bo Zhou: chaysezhou@tencent.com.},\; Fengzong Lian$^1$, Zhanhui Kang$^1$\\
\vspace{2mm}
\textbf{$^1$LLM Department, Hunyuan T1 Team, Tencent} \quad \textbf{$^2$The Chinese University of Hong Kong}
}

\begin{document}
\maketitle
\let\oldthefootnote\thefootnote

\let\thefootnote\oldthefootnote

\begin{abstract}
Recent advances in large language models (LLMs) have accelerated progress toward artificial general intelligence, with inference-time scaling emerging as a key technique. Contemporary approaches leverage either sequential reasoning (iteratively extending chains of thought) or parallel reasoning (generating multiple solutions simultaneously) to scale inference. However,
both paradigms face fundamental limitations: sequential scaling typically relies on arbitrary token budgets for termination, leading to inefficiency or premature cutoff; while parallel scaling often lacks coordination among parallel branches and requires intrusive fine-tuning to perform effectively. In light of these challenges, we aim to design a flexible test-time collaborative inference framework that exploits the complementary strengths of both sequential and parallel reasoning paradigms. Towards this goal, the core challenge lies in developing an efficient and accurate intrinsic quality metric to assess model responses during collaborative inference, enabling dynamic control and early termination of the reasoning trace.
To address this challenge, we introduce semantic entropy (SE), which quantifies the semantic diversity of parallel model responses and serves as a robust indicator of reasoning quality due to its strong negative correlation with accuracy. Building on this insight, we propose SEAT, a universal, plug-and-play framework that synergizes parallel exploration and sequential refinement by leveraging SE for adaptive termination. SEAT dynamically adjusts the degree of parallelization and employs SE-based stopping criteria, either via a statistical threshold or a novel threshold-free mechanism inspired by optimal stopping theory, to maximize efficiency without sacrificing performance.  Comprehensive evaluations on five challenging benchmarks demonstrate substantial accuracy gains, with as few as two parallel branches. Notably, SEAT also prevents catastrophic SE collapse in small-scale models during parallel scaling, thus maintaining robust performance.
\end{abstract}

\begin{figure}[htbp]
    \centering
    \vspace{-3mm}
    \includegraphics[width=0.9\linewidth]{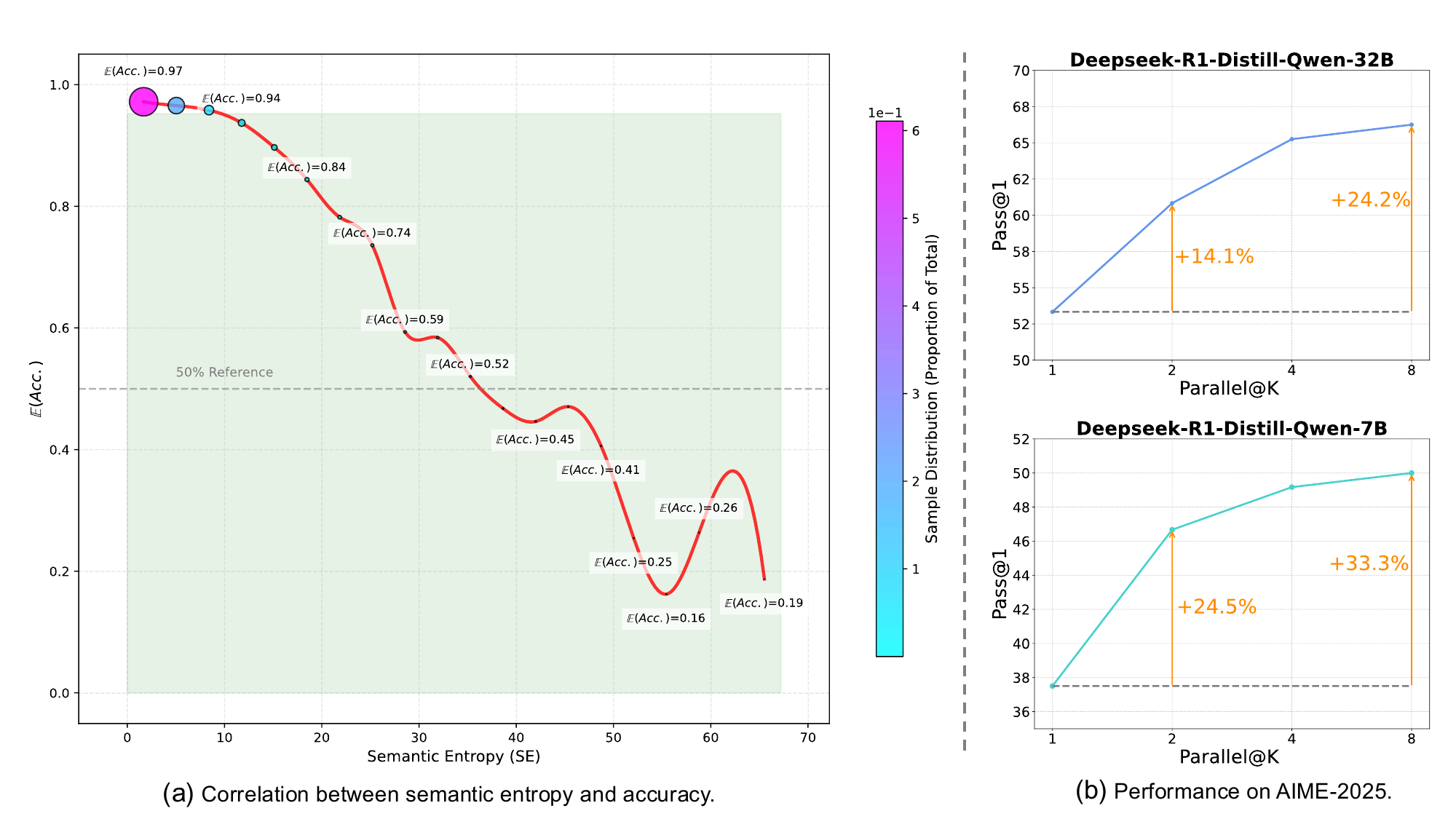}
    \caption{(a) The key insight of this paper: Strong negative correlation between semantic entropy and R1-Distill-Qwen-7B performance on Math-500 benchmark under $N=8$ parallel inferences per sample. The $\mathbb{E}(Acc.)$ denotes the expected accuracy and is calculated as the proportion of correct responses per $N$-inference set. (b) The performance of SEAT on R1-Distill-Qwen-7B/32B (AIME-2025), showing that our method yields substantial improvements.}
    \label{fig:enter-label}
\end{figure}

% \clearpage
\section{Introduction}
Recent advances in large language models (LLMs), exemplified by models such as o1~\citep{o1}, DeepSeek-R1~\citep{DeepSeekAI2025DeepSeekR1IR}, and QwQ~\citep{qwq}, have significantly accelerated progress toward artificial general intelligence. A key driver of this progress is test-time scaling~\citep{Snell2025ScalingLT, Welleck2024FromDT}, which improves performance by allowing the model to engage in more in-depth reasoning before producing a final answer, often reflected in the generation of more tokens with greater computational budget. Building on this insight, one line of research explores sequential scaling by introducing multi-round prompting for iterative refinement~\citep{Tian2025ThinkTE} or by explicitly steering the generation process through special tokens (e.g., ``wait’’) to encourage more deliberate reasoning. However, such sequential methods face an inherent constraint: in the absence of external supervision, the reasoning process is terminated based on fixed token budgets, often resulting in either unnecessary verbosity or premature halting \citep{chen2024not, wang2025thoughts}. In contrast, parallel scaling methods, such as best-of-N sampling~\citep{Cobbe2021TrainingVT, Kang2025ScalableBS} and majority voting~\citep{Wang2022SelfConsistencyIC}, promote diverse exploration through independent sampling but typically lack coordination among parallel inference paths. In this work, we investigate the following research question: \textbf{Can we design a flexible framework that effectively integrates the complementary advantages of both sequential and parallel scaling paradigms?}

The primary challenge in achieving this synergy lies in adaptively controlling the sequential scaling process for iterative refinement. This work addresses this challenge by drawing insight from parallel scaling. Intuitively, the uncertainty reflected by multiple independent samples can serve as a proxy for model performance \citep{Chen2023UniversalSF, Liang2024InternalCA, Zeng2025RevisitingTT}, as high semantic diversity among responses to the same prompt often indicates that the query exceeds the model’s current reasoning capability. In such cases, continuing the reasoning process until uncertainty drops below a predefined threshold tends to improve answer quality. Motivated by this intuition, we adopt semantic entropy (SE) \citep{Malinin2021UncertaintyEI, Chen2025SEEDGRPOSE}, which quantifies semantic diversity across multiple responses, as an intrinsic indicator of reasoning quality. We analyze the relationship between SE and answer quality, presenting experimental results on DeepSeek-R1-Distill-Qwen-7B \citep{DeepSeekAI2025DeepSeekR1IR} in Fig.~\ref{fig:enter-label}. A clear pattern emerges on the AIME-2025 benchmark, where model accuracy shows a strong negative correlation with semantic entropy. As SE increases, model accuracy steadily declines from $97\%$ to $19\%$. These findings demonstrate that \textbf{semantic entropy provides an effective and practical signal for monitoring and controlling the reasoning process, enabling the synergy between parallel and sequential scaling}.

\begin{figure}[!t]  % 允许 here / top / bottom / 新页    
\centering
    \includegraphics[width=0.75\linewidth]{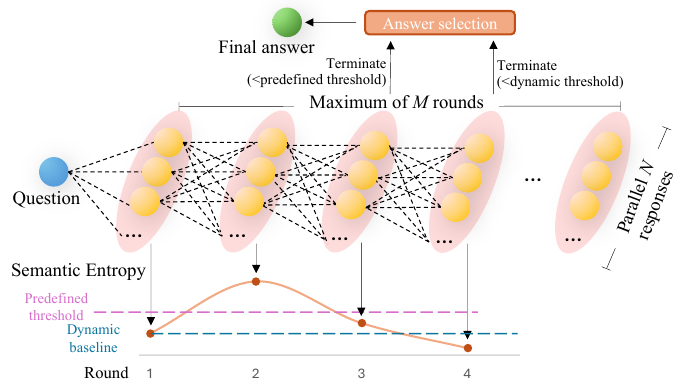}% 
    \caption{The overview of our proposed SEAT.}    
\label{fig:abstract}
\end{figure}

Based on the above insights, we propose SEAT\footnote{SEAT stands for a \textbf{S}emantic \textbf{E}ntropy-Guided \textbf{A}daptive \textbf{T}ermination Framework for Multi-round Parallel Reasoning.}, a universal, training-free parallel-reasoning framework. As shown in Fig. \ref{fig:abstract}, SEAT combines the advantages of parallel and sequential reasoning without extra fine-tuning. For parallel scaling, SEAT dynamically expands exploration by adjusting the degree of parallelization, preventing models from getting stuck on single reasoning paths. For sequential steps, it uses semantic entropy measurement to trigger early stopping, reducing wasted computation. To find an appropriate stopping threshold of specific $N$, we randomly sample one thousand mathematical questions and then inference $1000\times N$ results. For $N$ results in each parallel branch, we calculate the SE and randomly select one solution to evaluate accuracy. Statistical analysis revealed a similar $80/20$ pattern~\citep{Wang2025BeyondT8}: 80\% of correct answers had been selected from the lowest 20\% of the SE distribution. Therefore, we defined the threshold using the SE value at the 20th percentile. To further enable adaptive, threshold-free operation for eliminating computational overhead from pre-sampling, we draw inspiration from the mathematical resemblance between this problem and the ``Secretary Problem~\citep{Ferguson1989WhoST}'' (a classic case in Optimal Stopping Theory~\citep{Shiryaev1980OptimalSR}). Motivated by this connection, we develop an adaptive termination mechanism that establishes a semantic entropy baseline from the model’s initial reasoning steps and immediately stops the process when subsequent entropy readings fall below this baseline, thus requiring no predefined thresholds. 

We conduct comprehensive evaluation across five challenging benchmarks (AIME-2024, AIME-2025, MATH-500 \citep{Hendrycks2021MeasuringMP}, MINERVA \citep{lewkowycz2022solving}, GPQA~\citep{Rein2023GPQAAG}) using 7B and 32B models. Experimental results show that our adaptive method substantially improves model performance across all the five benchmarks. Notably, even with random answer selection, using the settings of as few as $N=2$ parallel responses, the 32B and 7B models attained remarkable accuracy gains of $14.1\%$ and $24.5\%$ on AIME-2025 respectively. Additionally, our framework demonstrates strong extensibility, i.e., seamlessly combining with max probability (i.e., the predicted probability of the response given by the LLM) and majority voting can further boost the performance. Surprisingly, we discovered that our adaptive approach effectively prevents semantic entropy collapse for smaller 7B models, a notorious phenomenon in which a sudden crash in SE occurs after certain sequential steps of parallelization. The reasoning process incurring semantic entropy collapse gets trapped in over-confidently repeating the same incorrect answers. Our adaptive termination method avoids this problem by concluding the reasoning process before entropy collapse happens, thereby enabling smaller 7B models to perform well with multi-round parallelization.

Our contribution can be summarized in threefold ways:
\begin{itemize}
    \item Revealing Negative Correlation: We establish a steep negative correlation between semantic entropy (SE) and a model's parallel reasoning performance, empirically showing that higher SE often accompanies lower accuracy (e.g., 97\% → 19\% accuracy drop on AIME2025 as SE increases). This suggests SE as a robust indicator for quality assessment in parallel reasoning.
    \item The SEAT Framework: We propose the SEAT framework, a semantic entropy (SE)-guided adaptive termination framework that synergizes parallel exploration and sequential refinement. Building on the observed 20/80 phenomenon,
    %(where 80\% of correct answers reside in the lowest 20\% SE quantile), 
    we first design a threshold-based termination strategy. Subsequently, inspired by optimal stopping theory, we develop a threshold-free mechanism that dynamically sets SE baselines during inference. Comprehensive evaluations across five benchmarks demonstrate significant performance gains of our proposed framework. 
    %Notably, even with a smaller parallelism (N=2), SEAT achieves significant improvements, yielding $14.7\%$ and $25.6\%$ accuracy gains for 32B and 7B models on AIME-2025.
    \item Discovering SE Collapse: We observe a critical phenomenon where semantic entropy (SE) collapses catastrophically in small-scale models (e.g., with $\leq 7 $ billion parameters) during parallel scaling with sequential steps. SEAT effectively avoid this collapse via early termination, sustaining stable performance.
\end{itemize}

\section{Methodology}
\subsection{Method Overview}
Given a question, this paper employs multi-round parallel reasoning to generate a collection of candidate responses, and then selects the final answer from this set. Three primary components are involved in the above procedure: (1) the design of the multi-round parallel reasoning framework, (2) calculation of the SE metric during inference with termination mechanism, and (3) selection strategy for the final answer among candidate responses. These components will be described step by step.

\paragraph{Multi-round Parallel (MRP) Inference Framework:}
As illustrated in Fig.~\ref{fig:abstract}, the proposed SEAT framework establishes an $N\times M$ reasoning structure for multi-round parallel reasoning, where $N$ represents the parallel dimension (\textit{i.e.}, number of reasoning paths per round) and $M$ is the sequential dimension (\textit{i.e.}, number of reasoning rounds). Notably, $N\times M$ serves as the pre-defined maximum reasoning budget allocated to the model. Due to our proposed adaptive termination mechanism during inference, the actual computational budget typically falls below this pre-defined maximum. Details on this mechanism will be described in Sec. \ref{sec:experiments}. In contrast to prior test-time scaling approaches, SEAT synergistically integrates sequential refinement and parallel exploration. Specifically, each sequential round can access all $N$ reasoning outputs from the prior round, enabling the model to refine its reasoning by leveraging diverse responses for error correction. Furthermore, within each round, all parallel reasoning paths operate independently, which aims to increase the diversity and encourage exploration. Formally, the $j$-th reasoning path in the $i$-th round is defined as:
\begin{equation}
    MRP^i_j(P_{i}) \rightarrow \{\text{Thinking}^{i}_{j}, \text{Answer}^{i}_{j}\},
\end{equation}
where $i\in\{1, \cdots, M\}$ and $j\in\{1, \cdots, N\}$. Here, $P_{i}$ denotes the input prompt for the $i$-th round, which is initialized as the user prompt (i.e., $P_1=\langle\text{user prompt}\rangle$). For every round from the second onward, we extract the answer segments of all $N$ candidate responses produced by the parallel reasoning paths in the preceding round, and incorporate these $N$ answers into the prompt for the current round. The prompt template for $P_i$ in the $i$-th round ($i \in \{2, \ldots, M\}$) is as follows:

\begin{AIbox}{Prompt Template for the $i$-th Round ($i \geq 2$)}
Original question prompt:\newline
\textless User prompt \textgreater
\newline
-{}-{}-{}-{}-{}-{}-{}-{}-{}-{}-\newline
I will provide several assistant’s previous answers, please analyze and then re-answer.\newline

\#\#\# 1. The assistant’s previous answer is: \textless $\text{Answer}^{i-1}_{1}$ \textgreater
\newline
... 
\newline
\#\#\# $N$. The assistant’s previous answer is: \textless $\text{Answer}^{i-1}_{N}$ \textgreater

-{}-{}-{}-{}-{}-{}-{}-{}-{}-{}-\newline
The above are the questions along with the assistant's previous answers, please analyze and then re-answer.
\end{AIbox}
% It can be seen that the prompt construction for the $i$-th round depends solely on the answer from the prior round ($i-1$). This design encourages the model to comprehensively analyze prior results and generate improved responses.
By constructing the prompt for the $i$-th round using $N$ parallel answers generated in the previous round, this approach encourages the model to review and refine its previous outputs for improved responses.

\paragraph{Semantic Entropy Calculation and Termination Mechanism:} 
% Given a user prompt $q$, and a collection of $N$ parallel inference responses, we now describe the method for computing semantic entropy.
% Given a question $q$ and the $N$ answers from the previous round $\{a_1, \ldots, a_N\}$, 
Given a question $q$ and $N$ answers $\{a_1, \ldots, a_N\}$ extracted from the previous round’s $N$ responses, we compute the semantic entropy to quantify the model’s uncertainty about $q$ in light of these answers. Since the semantic entropy computation process is applied across different reasoning rounds, we omit the round-indicating superscripts for simplicity. The semantic entropy is defined as follows:
\begin{equation} \label{equ:se1}
    \mathrm{SE}(q) = -\sum_{c} \left( \left( \sum_{r \in c} p\left( r \mid \{q; a_1, \ldots, a_N \} \right) \right) \log \left[ \sum_{r \in c} p\left( r \mid \{q; a_1, \ldots, a_N \} \right) \right] \right),
\end{equation}
where $c$ denotes a possible semantic meaning class and $r$ denotes a possible response. It's intractable to enumerate every possible $c$ since LLMs can generate an unlimited number of diverse responses for a given question, potentially spanning numerous unknown semantic categories. To this end, we estimate \eqref{equ:se1} using  Monte Carlo approximation~\citep{Kuhn2023SemanticUL,Farquhar2024DetectingHI}:
\begin{equation} \label{equ2}
    \mathrm{SE}(q) \approx -{|\mathcal{C}|}^{-1}\sum^{|\mathcal{C}|}_{k=1}\log p \left(\mathcal{C}_k \mid \{q;a_1, \ldots, a_N\} \right),
\end{equation}
where $\mathcal{C}=\{\mathcal{C}_1, \mathcal{C}_2, \cdots, \mathcal{C}_K\}$ denotes the semantic cluster. In practice, we sample $N$ responses $\{r_1, \ldots, r_N\}$ and perform clustering based solely on the final answer segment of each $r_i$ to obtain the set of semantic classes $\mathcal{C}$. The cluster probability in \eqref{equ2} is subsequently calculated as:
\begin{equation}
    p(\mathcal{C}_k \mid \{q;a_i, \ldots, a_N\} ) = \sum_{i=1}^{N} \mathbb{I}[r_i \in \mathcal{C}_k]\, p(r_i \mid \{q;a_i, \ldots, a_N\}).
\end{equation}
It's worth noting that, since the thinking part of $r_i$ can often span tens of thousands of tokens, we compute only the probability of the answer part in practical computation\footnote{This leads to inflated probability estimates, occasionally resulting in negative SE values. However, since these occurrences do not invalidate the observed negative correlation between SE and model performance, we do not apply corrections to the metric.}.  

Based on the obtained SE for the response generated in round $i$, we design a mechanism to determine whether to terminate the reasoning process. Specifically, leveraging the clear inverse correlation observed between SE and performance where higher SE signals greater uncertainty and lower response quality, we dynamically control the reasoning process by monitoring SE after each iteration. The reasoning continues into the next round if the measured SE exceeds a set threshold, implying the output requires further refinement, and is terminated only once the SE meets a pre-defined stopping condition. The specific configuration of this stopping condition is detailed in Sec. \ref{sec:atwpt} and Sec. \ref{sec:dtfmost}.

\paragraph{Answer Selection Strategy.} 
Once the reasoning process has terminated, we need to select the final response from the candidate answers. Without recourse to external validators, we employ three conventional strategies: random selection, maximum probability selection, and majority voting; the effectiveness of these approaches will be thoroughly examined in our experiments. Additionally, a worthy question is how to define the candidate answer pool. In this paper, instead of collecting all responses across the previous round, we propose to exclusively utilize responses generated in the terminal reasoning round as the candidate set. This design is motivated by our observation that the preceding output is likely to be suboptimal due to the higher elevated uncertainty. Incorporating such under-refined responses would introduce detrimental noise, thereby degrading model performance. 
%The empirical impact will be further analyzed in the experimental results.

\begin{figure}[!t]
    \centering
    % 直接并列插入图片，并手动控制间距
    \includegraphics[width=0.3\linewidth]{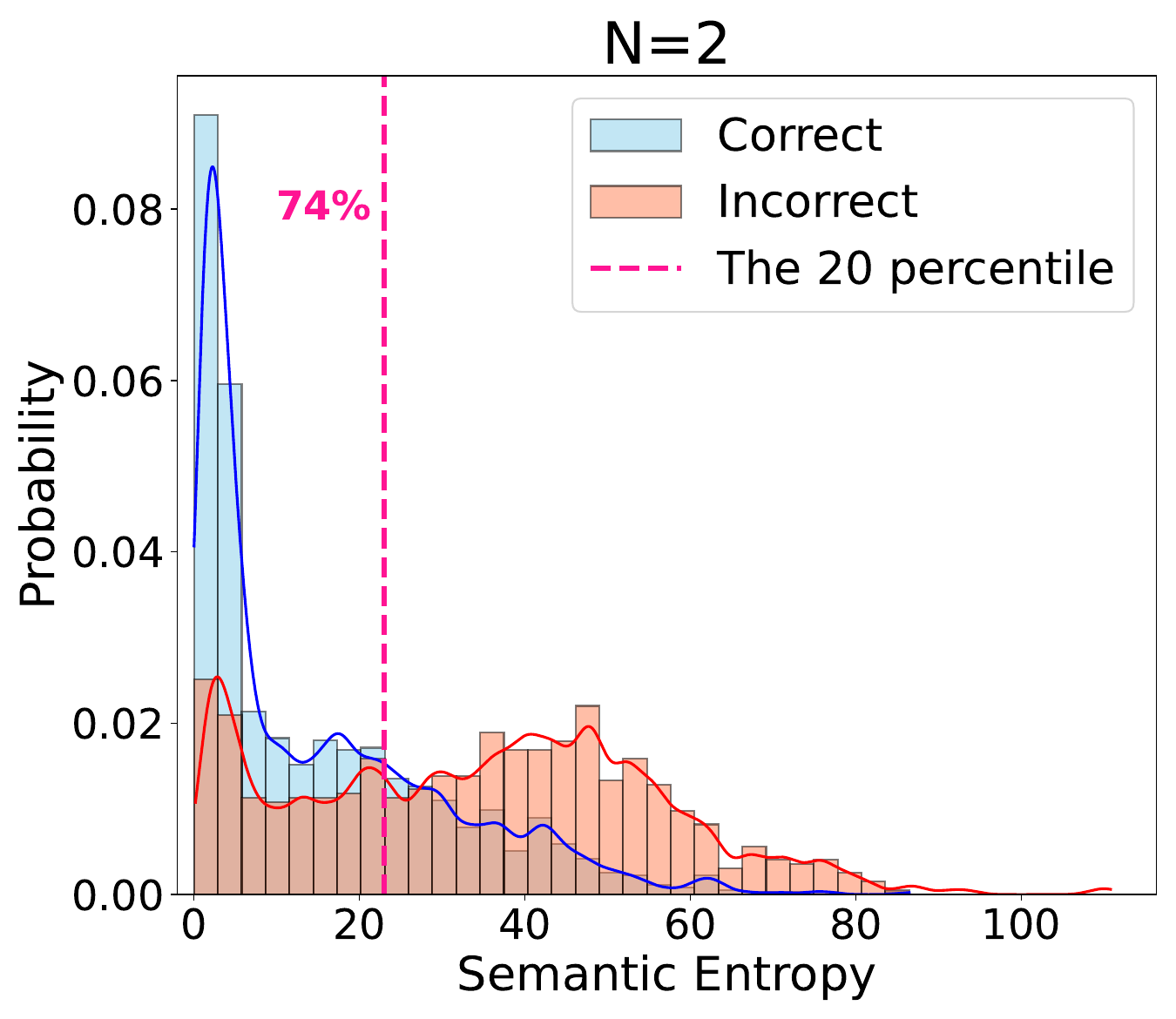}
    \includegraphics[width=0.3\linewidth]{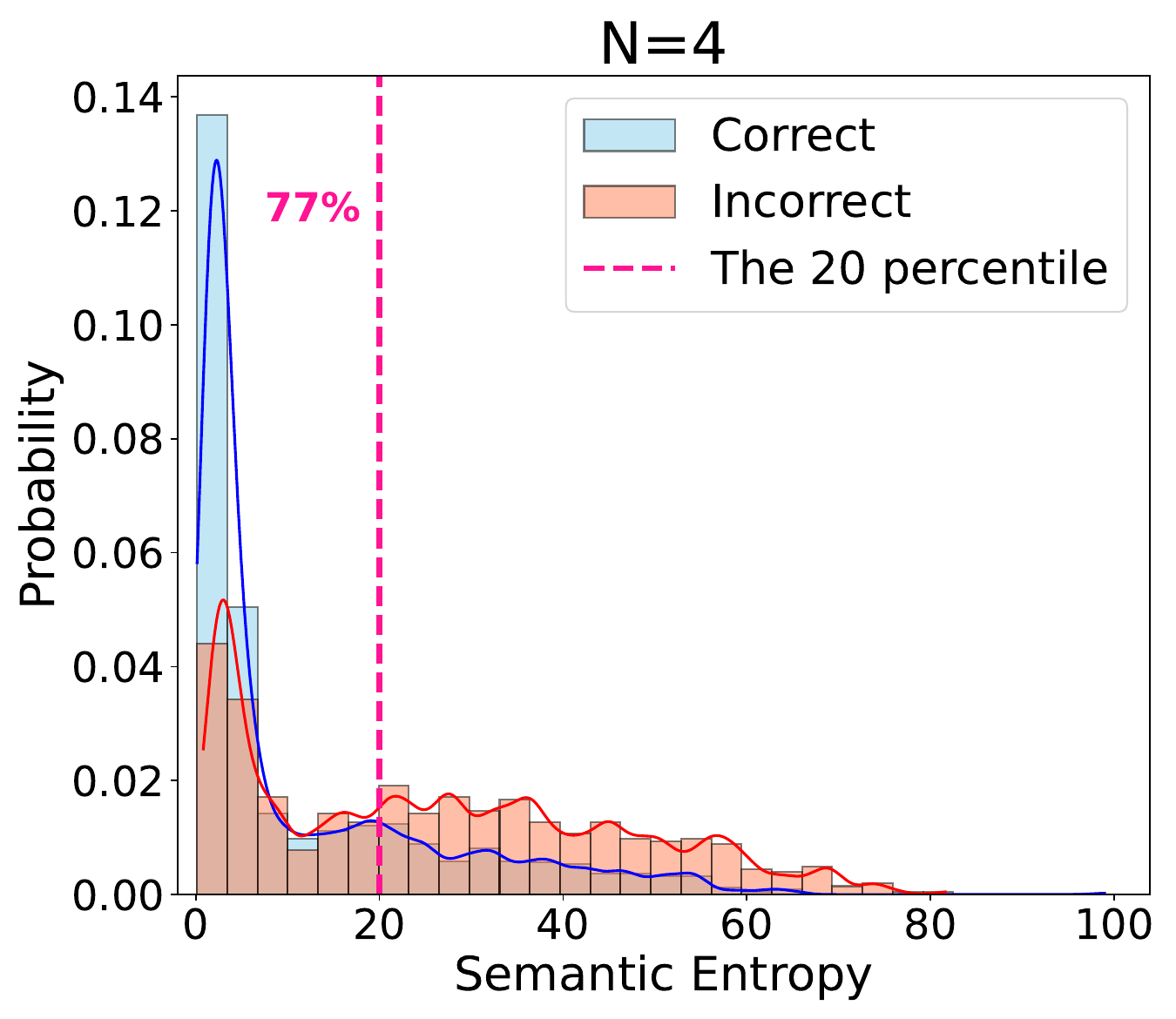}
    \includegraphics[width=0.3\linewidth]{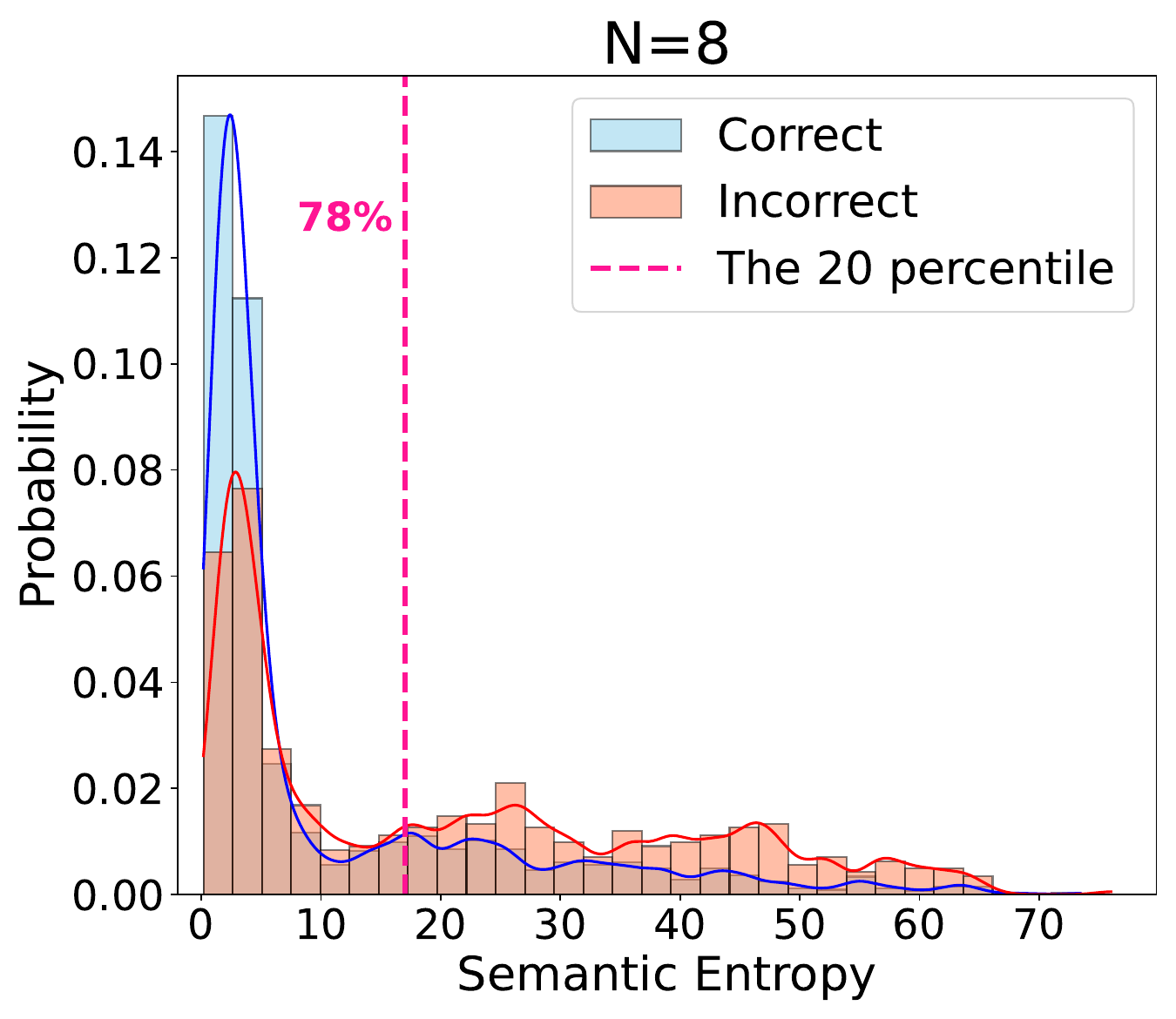}
    
    % 整个图片组合的总标题
    \caption{Semantic entropy distribution of correct and incorrect answers. The lowest 20\% threshold is marked by red line and the proportion of correct answers within this region is labeled numerically in red.}
    \label{fig:method_80_20}
\end{figure}

\subsection{Adaptive Termination with Pre-defined Threshold} \label{sec:atwpt}
This section details the procedure for establishing reasoning termination conditions. A widely adopted approach is to establish a pre-defined SE threshold, halting the inference procedure when monitored SE is lower than this calibrated value. To this end, we first conducted foundational experiments using DeepSeek-R1-Distill-Qwen-7B model on our proprietary dataset of $1000$ challenging mathematical problems. For each problem, we randomly selected one candidate solution from parallel inference outputs for evaluation. Then we analyzed the SE distribution of correct and incorrect answers, and the statistical analysis is visualized in Fig.~\ref{fig:method_80_20}. The statistical analysis reveals an $80/20$ pattern wherein about $80\%$ of the correct answers lie in the lowest quintile of the SE distribution. Specifically, for $N=2$, approximately $74\%$ of correct answers fall within the specified range, increasing to $77\%$ at $N=4$ and reaching $78\%$ for $N=8$. Leveraging this pattern, given the model and a specific parallel degree, we can first sample parallel responses on substantial data. Then, we compute the empirical SE distribution and select the 20th-percentile value as the pre-defined threshold.

\begin{table}[!t]
\centering
\begin{tabular}{c|c|c|c|c|c|c}
\toprule
 \multicolumn{2}{c|}{DeepSeek-R1-Distill-Qwen-32B} & \textbf{AIME-2024} & \textbf{AIME-2025} & \textbf{MATH-500} & \textbf{MINERVA} & \textbf{GPQA} \\
\midrule
\multicolumn{7}{c}{Random} \\
\midrule
\multirow{4}{*}{n=2} & baseline & 70.83 & 53.33 & 95.12 & 57.36 & 66.87 \\
 & Ours (Fixed) & 75.42 & 56.25 & 95.45 & 58.50 & 67.93\\
 & Ours (Adaptive) & 78.75 & 60.83 & 96.05 & 58.72 & 68.06 \\
 & Ours (Min) & \textbf{82.35} & \textbf{65.83} & \textbf{96.20} & \textbf{58.78} & \textbf{68.12} \\
\midrule
\multirow{4}{*}{n=4}  & baseline & 70.83 & 53.45 & 95.11 & 57.14 & 66.54 \\ 
 & Ours (Fixed) & 78.33 & 64.17 & 96.03 & \textbf{59.28} & 67.23\\
 & Ours (Adaptive) & 80.41 & 65.25 & 96.38 & 58.51 & 67.42 \\
 & Ours (Min) & \textbf{83.33} & \textbf{70.83} & \textbf{96.65} & 57.58 & \textbf{67.80} \\
 \midrule
\multirow{4}{*}{n=8}  & baseline & 70.42 & 52.92 & 95.13 & 57.85 & 65.85 \\
 & Ours (Fixed) & 80.42 & 65.83 & 96.53 & \textbf{59.56} & 66.60 \\
 & Ours (Adaptive) & 85.67 & 66.25 & 96.85 & 58.36 & 68.57 \\
 & Ours (Min) & \textbf{85.83} & \textbf{69.16} & \textbf{96.98} & 58.55 & \textbf{68.88} \\
\midrule 
\multicolumn{7}{c}{Max Probability} \\
\midrule
\multirow{4}{*}{n=2} & baseline & 72.92 & 53.73 & 95.22 & 58.01 & 66.74 \\
 & Ours (Fixed) & 74.17 & 55.00 & 95.43 & 58.87 & 67.74 \\
 & Ours (Adaptive) & 79.32 & 60.83 & 96.25 & 59.13 & 68.12 \\
 & Ours (Min) & \textbf{82.50} & \textbf{65.83} & \textbf{96.25} & \textbf{59.45} & \textbf{68.12} \\
\midrule
\multirow{4}{*}{n=4} & baseline & 72.50 & 55.83 & 95.54 & 57.00 & 66.54 \\ 
 & Ours (Fixed) & 75.83 & 59.17 & 95.98 & 58.60 & 66.67 \\
 & Ours (Adaptive) & 81.67 & 67.08 & \textbf{96.70} & \textbf{59.88} & 67.42 \\
 & Ours (Min) & \textbf{83.33} & \textbf{71.25} & 96.65 & 57.67 & \textbf{67.42} \\
 \midrule
\multirow{4}{*}{n=8} & baseline & 70.83 & 55.00 & 95.33 & 57.58 & 65.66 \\ 
 & Ours (Fixed) & 79.17 & 64.17 & 95.78 & 58.04 & 66.35 \\
 & Ours (Adaptive) & 85.87 & 67.08 & 96.75 & 58.86 & 68.87 \\
 & Ours (Min) & \textbf{86.25} & \textbf{71.25} & \textbf{96.98} & \textbf{59.05} & \textbf{69.13} \\
\midrule 
\multicolumn{7}{c}{Majority Voting} \\
\midrule
\multirow{4}{*}{n=2} & baseline & 72.92 & 53.75 & 95.12 & 58.19 & 67.68  \\
 & Ours (Fixed) & 74.17 & 55.00 & 95.45 & 58.95 & 67.87 \\
 & Ours (Adaptive) & 80.00 & 61.67 & 96.30 & 59.35 & 68.06 \\
 & Ours (Min) & \textbf{82.50} & \textbf{65.83} & \textbf{96.23} & \textbf{59.45} & \textbf{68.08} \\
\midrule
\multirow{4}{*}{n=4} & baseline & 80.83 & 64.17 & 95.84 & 58.47 & 67.84 \\ 
 & Ours (Fixed) & 81.43 & 66.67 & \textbf{96.80} & \textbf{60.07} & \textbf{68.24} \\
 & Ours (Adaptive) & 82.50 & 68.33 & 96.53 & 59.01 & 68.12 \\
 & Ours (Min) & \textbf{83.33} & \textbf{70.83} & 96.67 & 57.76 & 67.55 \\
 \midrule
\multirow{4}{*}{n=8} & baseline & 83.75 & 65.83 & 96.55 & 59.05 & 68.32 \\
 & Ours (Fixed) & 85.83 & 70.42 & 96.98 & \textbf{59.78} & \textbf{69.44}\\
 & Ours (Adaptive) & 85.83 & 69.17 & \textbf{97.08} & 59.36 & 69.32 \\
 & Ours (Min) & \textbf{85.83} & \textbf{70.83} & 97.00 & 58.69 & 68.94 \\
\bottomrule
\end{tabular}
\caption{Multi-round parallel reasoning performance of DeepSeek-R1-Distill-Qwen-32B on different datasets.}
\label{fig:32b_main_result}
\end{table}

\begin{table}[!t]
\centering
\begin{tabular}{c|c|c|c|c|c|c}
\toprule
 \multicolumn{2}{c|}{DeepSeek-R1-Distill-Qwen-7B} & \textbf{AIME-2024} & \textbf{AIME-2025} & \textbf{MATH-500} & \textbf{MINERVA} & \textbf{GPQA} \\
\midrule
\multicolumn{7}{c}{Random} \\
\midrule
\multirow{4}{*}{n=2} & baseline & 60.41 & 37.50 & 93.95 & 52.14 & 53.09 \\
 & Ours (Fixed) & 65.00 & 47.08 & 94.93 & 54.27 & 55.05 \\
 & Ours (Adaptive) & 64.58 & 46.67 & 95.28 & 53.54 & 55.11 \\
 & Ours (Min) & \textbf{66.67} & \textbf{51.25} & \textbf{95.20} & \textbf{54.83} & \textbf{55.68} \\
\midrule
\multirow{4}{*}{n=4} & baseline & 58.33 & 42.92 & 94.03 & 52.58 & 51.58 \\
 & Ours (Fixed) & 68.75 & 46.25 & 95.03 & \textbf{56.57} & 55.18 \\
 & Ours (Adaptive) & 68.33 & 48.17 & 95.50 & 55.97 & 55.47 \\
 & Ours (Min) & \textbf{70.83} & \textbf{48.75} & \textbf{95.80} & 54.64 & \textbf{56.25} \\
 \midrule
\multirow{4}{*}{n=8} & baseline & 56.25 & 40.83 & 93.78 & 52.80 & 53.09 \\
 & Ours (Fixed) & 68.75 & 50.00 & 95.45 & 55.38 & \textbf{57.95} \\
 & Ours (Adaptive) & 68.33 & 50.00 & 95.35 & \textbf{55.71} & \textbf{57.95} \\
 & Ours (Min) & \textbf{71.67} & \textbf{51.25} & \textbf{95.53} & 55.00 & 57.07 \\
\midrule 
\multicolumn{7}{c}{Max Probability} \\
\midrule
\multirow{4}{*}{n=2} & baseline & 64.17 & 41.67 & 94.37 & 53.08 & 54.37 \\
 & Ours (Fixed) & 65.00 & 46.67 & 95.03 & 53.77 & \textbf{55.81} \\
 & Ours (Adaptive) & 65.42 & 46.25 & \textbf{95.35} & 54.78 & 55.18 \\
 & Ours (Min) & \textbf{65.83} & \textbf{51.67} & 95.28 & \textbf{55.10} & 55.43 \\
\midrule
\multirow{4}{*}{n=4} & baseline & 66.67 & 44.58 & 94.78 & 54.61 & 54.73 \\ 
 & Ours (Fixed) & 70.42 & 48.75 & 95.60 & 55.28 & 55.05 \\
 & Ours (Adaptive) & 70.00 & 50.00 & \textbf{95.88} & \textbf{55.84} & 56.00 \\
 & Ours (Min) & \textbf{71.67} & \textbf{52.08} & 95.70 & 55.10 & \textbf{56.44} \\
 \midrule
\multirow{4}{*}{n=8} & baseline & 64.58 & 45.42 & 94.55 & 54.87 & 56.07 \\
 & Ours (Fixed) & 68.75 & 50.42 & 95.68 & 55.42 & \textbf{58.84} \\
 & Ours (Adaptive) & 68.75 & 50.83 & \textbf{95.95} & \textbf{56.27} & \textbf{58.84} \\
 & Ours (Min) & \textbf{70.42} & \textbf{51.67} & 95.60 & 55.23 & 57.95 \\
\midrule 
\multicolumn{7}{c}{Majority Voting} \\
\midrule
\multirow{4}{*}{n=2} & baseline & 63.75 & 41.25 & 94.48 & 53.45 & 55.19 \\
 & Ours (Fixed) & 65.00 & 46.25 & 95.00 & 54.04 & \textbf{55.74} \\
 & Ours (Adaptive) & \textbf{65.42} & 46.67 & \textbf{95.38} & \textbf{54.83} & 55.68 \\
 & Ours (Min) & \textbf{65.42} & \textbf{51.25} & 95.25 & 54.73 & 55.49 \\
\midrule
\multirow{4}{*}{n=4} & baseline & 70.33 & 48.58 & 95.80 & 55.02 & 55.78 \\ 
 & Ours (Fixed) & \textbf{72.08} & 49.58 & 95.83 & \textbf{55.79} & 56.19\\
 & Ours (Adaptive) & 71.67 & 49.58 & \textbf{95.93} & 55.10 & \textbf{56.44} \\
 & Ours (Min) & 70.83 & \textbf{52.08} & 95.83 & 55.00 & 56.19 \\
 \midrule
\multirow{4}{*}{n=8} & baseline & 71.42 & 52.83 & 95.35 & 56.02 & 57.09 \\
 & Ours (Fixed) & \textbf{72.92} & 52.92 & \textbf{95.98} & 56.42 & 58.84\\
 & Ours (Adaptive) & 72.38 & \textbf{55.00} & 95.88 & \textbf{56.74} & \textbf{58.95} \\
 & Ours (Min) & 70.83 & 52.08 & 95.60 & 56.19 & 57.51 \\
\bottomrule
\end{tabular}
\caption{Multi-round parallel reasoning performance of DeepSeek-R1-Distill-Qwen-7B on different datasets.}
\label{fig:7b_main_result}
\end{table}

\subsection{Adaptive Threshold-free Mechanism} \label{sec:dtfmost}
Although the threshold-based approach described above provides a principled stopping criterion, it requires extensive pre-sampling and recalibration when model configurations or parallel settings change. We consequently develop a more adaptive threshold-free method to avoid this shortcoming. To this end, we reformulate our goal to identify the optimal reasoning round exhibiting minimal SE under a fixed inference budget. Surprisingly, we found that this goal is quite similar to the ``Secretary Problem~\citep{Ferguson1989WhoST}''.  The classical secretary problem constitutes a foundational problem in Optimal Stopping Theory~\citep{Shiryaev1980OptimalSR}, addressing sequential selection under uncertainty. It aims to maximize the probability of selecting the best candidate from an unknown sequence of applicants when interviews must be conducted irreversibly without recall. The core strategy used in secretary problem establishes a qualification baseline by observing the first $T$  candidates during an initial exploration phase. Subsequent candidates are evaluated against this dynamically determined threshold, with immediate selection of the first applicant exceeding the baseline. This observation-then-selection framework aligns with our objective of identifying optimal termination round during multi-round reasoning. Drawing inspiration from this problem, we propose an adaptive threshold-free approach. Specifically, our method dynamically sets the adaptive threshold as the minimal SE observed during the initial $T$ rounds and terminates inference immediately when encountering any round with SE below this calibrated minimum. Given the computational expense of long CoT sampling, we set $T=1$, defining the dynamic threshold exclusively from the first reasoning round. This reduces computational overhead while maintaining empirically competitive performance.

\section{Experiments} \label{sec:experiments}
\subsection{Experimental Setup}
We evaluate the propsoed SEAT method on AIME-2024, AIME-2025, MATH-500\citep{Hendrycks2021MeasuringMP}, MINERVA, and GPQA~\citep{Rein2023GPQAAG} benchamrk. These benchmarks cover multiple domains and difficulty levels, allowing for a thorough assessment of reasoning performance across varied scenarios.

To ensure a fair and consistent comparison, we employed the publicly available DeepSeek-R1-Distill-Qwen-32B (shorted as R32B)\footnote{https://huggingface.co/deepseek-ai/DeepSeek-R1-Distill-Qwen-32B} and DeepSeek-R1-Distill-Qwen-7B (shorted as R7B)\footnote{https://huggingface.co/deepseek-ai/DeepSeek-R1-Distill-Qwen-7B}~\citep{DeepSeekAI2025DeepSeekR1IR} as the inference baseline models. For reproducibility, we maintained unified inference settings throughout all evaluations. The maximum generation length is set to 32768 tokens, while temperature and top-p are applied to 0.7 and 0.95, respectively. Each query across all datasets will be conducted 8 independent multi-round parallel inferences and reports the average accuracy. Although our methodology incorporates an adaptive termination mechanism, we processed every query through the full $N\times M$ inference procedure (where we set $M=8$ for max rounds) to enable detailed experimental analysis. 

To define an appropriate termination threshold, we systematically analyzed the SE distribution across randomly pre-sampled problem instances. We extracted 10000 challenging mathematical problems from our proprietary dataset and then conducted sampling based on varying model scales (7B and 32B) and $N\in \{2,4,8\}$ parallelization degree. Subsequently, the distribution of SE was analyzed to determine an empirical threshold defined as the 20th percentile value of the distribution\footnote{To guarantee the rationality of the distribution, we excluded roughly 5\% of outlier points within the long tail of the results.}. To mitigate selection bias and ensure practical utility, these percentile-derived thresholds were rounded to the nearest integer. Consequently, for the 32B model architecture, the finalized thresholds were set at 27 ($N=2$), 22 ($N=4$), and 19 ($N=8$). Corresponding thresholds for the 7B model were determined as 23 ($N=2$), 20 ($N=4$), and 17 ($N=8$).

\subsection{Main Results}

We evaluate the effectiveness of different components as follows: $\blacktriangleright$\textit{Fixed:} we used the pre-defined threshold for inference termination. $\blacktriangleright$\textit{Adaptive:} the adaptive threshold-free mechanism is applied. $\blacktriangleright$\textit{Min:} the inference round with minimal SE among $M$ round is selected. The experimental results on R32B and R7B are shown in Table~\ref{fig:32b_main_result} and Table~\ref{fig:7b_main_result}, respectively.

It can be seen that, our approach achieves substantial performance gains of both R32B and R7B models across different datasets. Specifically, under the random pick selection and adaptive setting, the R32B model achieves remarkable improvements including 70.83 to 85.67 (+21.0\%) on AIME-2024 and 53.33 to 66.25 (+24.2\%) on AIME-2025, along with average gains of +2.5\% across MATH-500, MINERVA, and GPQA. Similarly, the R7B model demonstrates consistent gains under identical settings, improving from 60.41 to 68.33 (+13.1\%) on AIME-2024 and from 37.50 to 50.00 (+33.3\%) on AIME-2025, while achieving +6.12\% average improvement across MATH-500, MINERVA, and GPQA. These results collectively confirm the effectiveness of our framework. Notably, even at minimal parallelization (N=2), our method delivers remarkable gains with the R32B model improving from 53.33 to 60.83 (+14.1\%) and the R7B model advancing from 37.50 to 46.67 (+24.5\%) on AIME-2025. Furthermore, integrating max probability and majority voting strategies yields additional performance gains, with R32B model showing 0.6\% (max probability) and 1.5\% (majority voting) average improvements and R7B model achieving 1.1\% and 4.0\% gains, respectively. This demonstrates the scalability of our proposed framework.

Meanwhile, our method approach using pre-defined thresholds significantly outperforms the baselines, demonstrating that pre-sampling probes of model SE distributions serve as a sound strategy. It is worth noting that although the thresholds were determined exclusively using mathematical datasets, it can also bring improvements on GPQA dataset, which demonstrates robust generalization capability of the proposed framework. Additionally, the adaptive threshold-free method usually surpasses fixed approaches. We hypothesize that this pattern indicates divergent SE distributions across different problem types for identical models. Establishing SE baseline from model's earlier reasoning outputs can better capture these dynamics, thereby yielding heightened performance and improved generalization capability. Surprisingly, minimum selection (min) sometimes degrades performance in the R7B model at higher $N$ values. For instance at $N=8$, the AIME-2025 score drops to 52.08 versus the baseline 52.83. We attribute this phenomenon to SE collapse during parallel inference in smaller models and provide comprehensive analysis in Sec. \ref{sec:VSC}.

\begin{figure}[!t] 
	\centering
	\subfigure[R32B under $N=2$ parallelization.]{
		\begin{minipage}[t]{0.49\linewidth}
			\centering
			\includegraphics[width=3.3in]{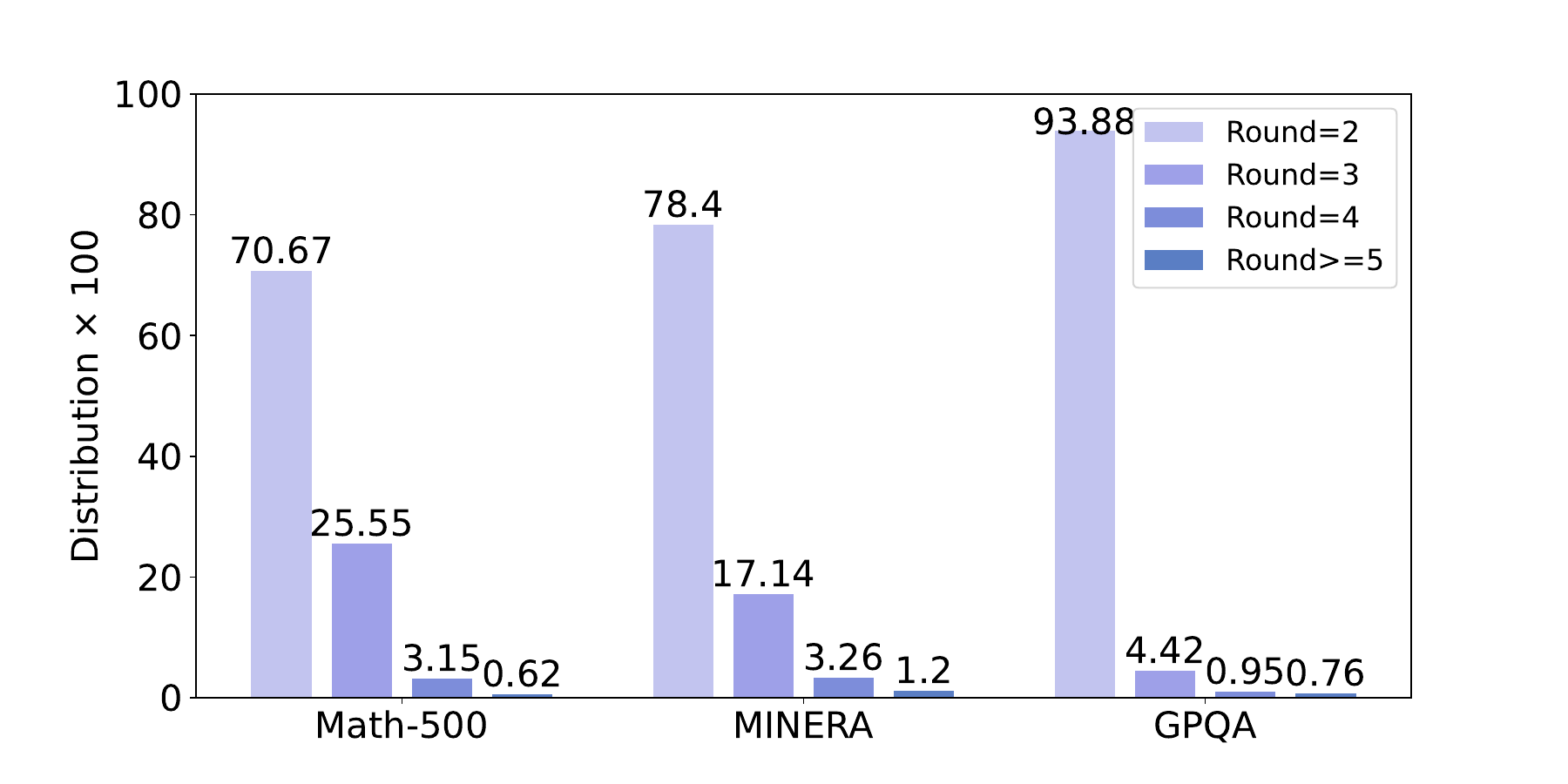}
			%\caption{fig2}
		\end{minipage}%
	}
        \subfigure[R32B under $N=8$ parallelization.]{ 
		\begin{minipage}[t]{0.49\linewidth}
			\centering
			\includegraphics[width=3.3in]{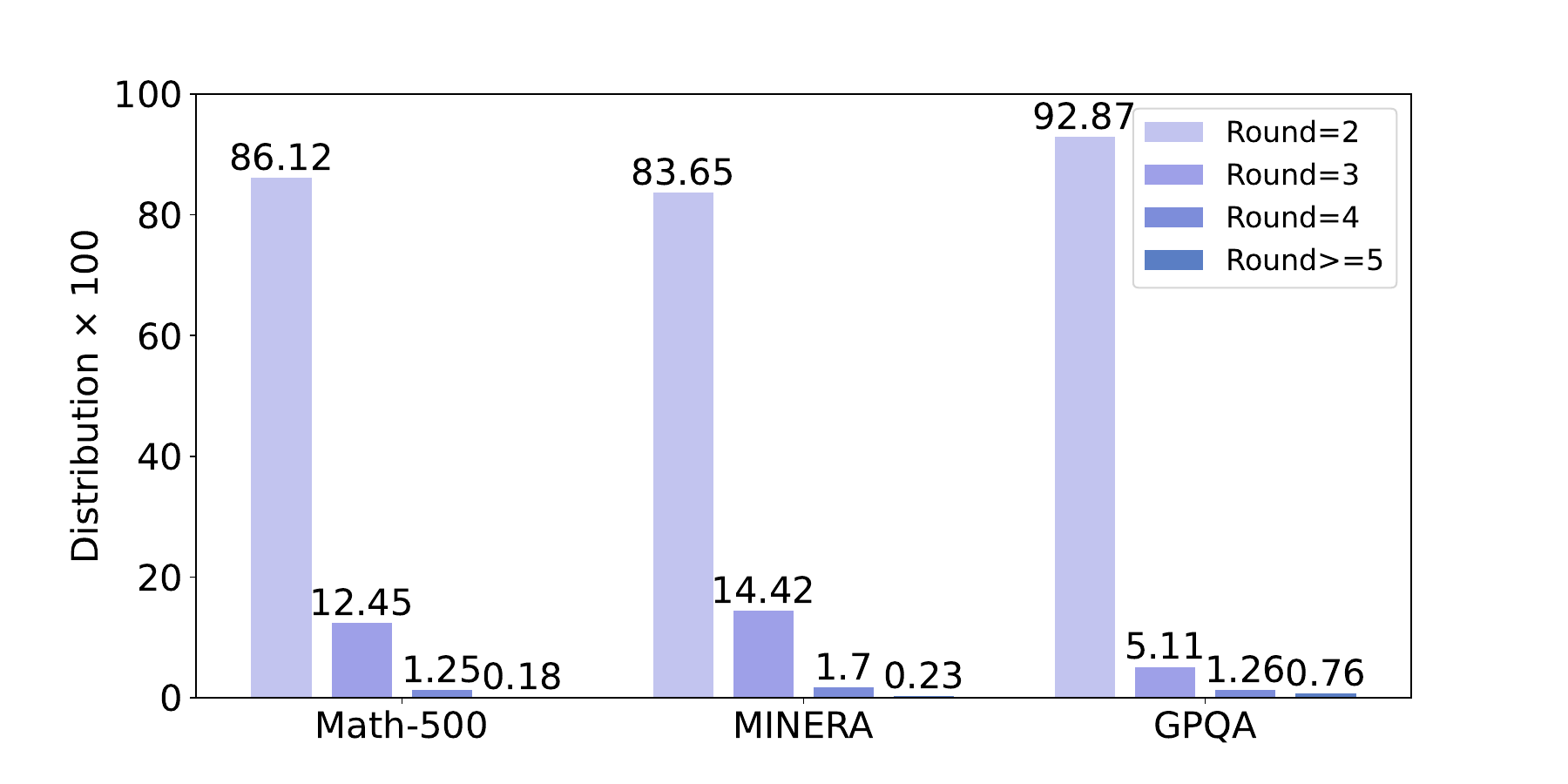}
			%\caption{fig1}
		\end{minipage}%
	}%
	
	\subfigure[R7B under $N=2$ parallelization.]{
		\begin{minipage}[t]{0.49\linewidth}
			\centering
			\includegraphics[width=3.3in]{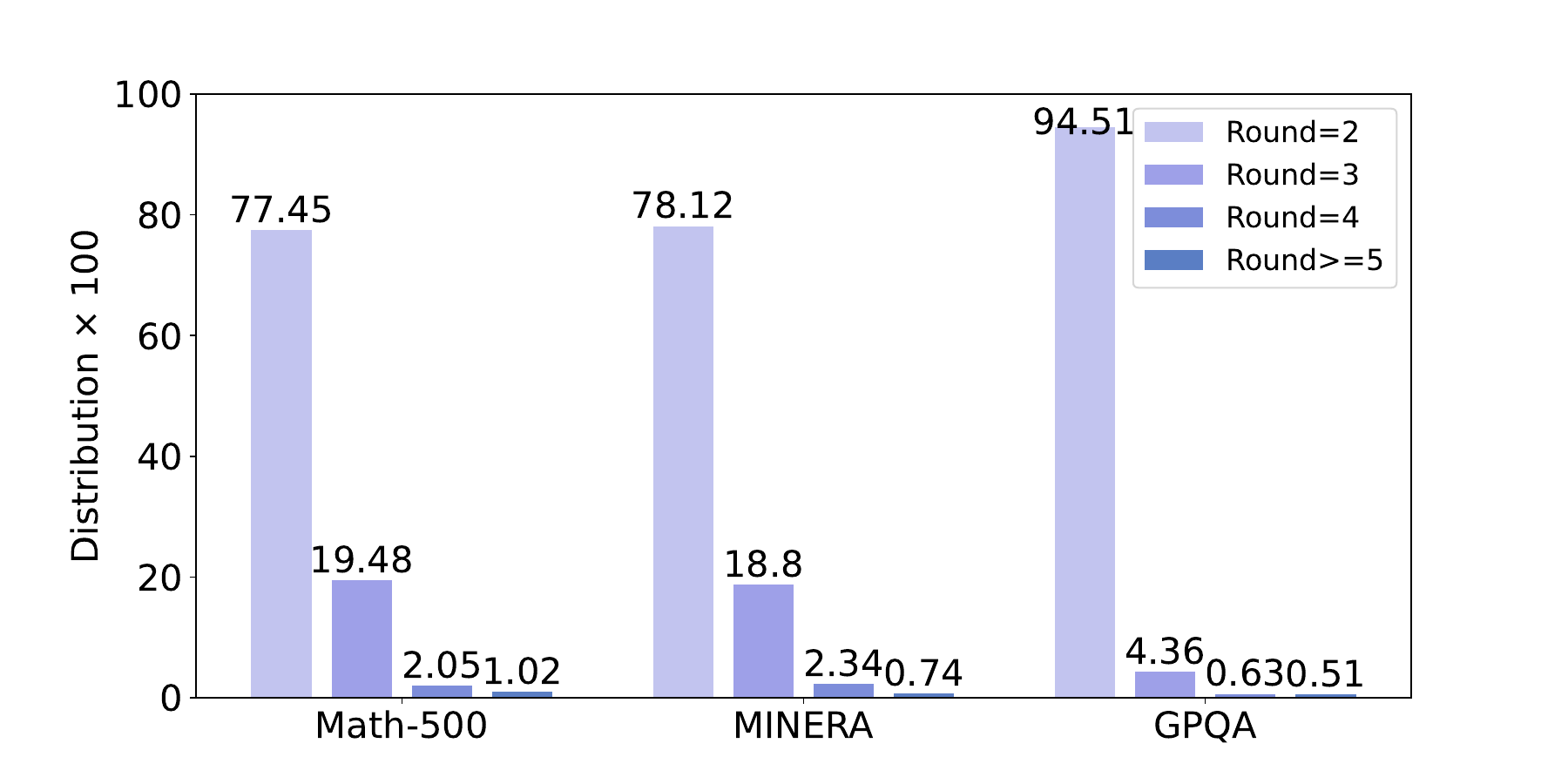}
			%\caption{fig2}
		\end{minipage}%
	}
        \subfigure[R7B under $N=8$ parallelization.]{ 
		\begin{minipage}[t]{0.49\linewidth}
			\centering
			\includegraphics[width=3.3in]{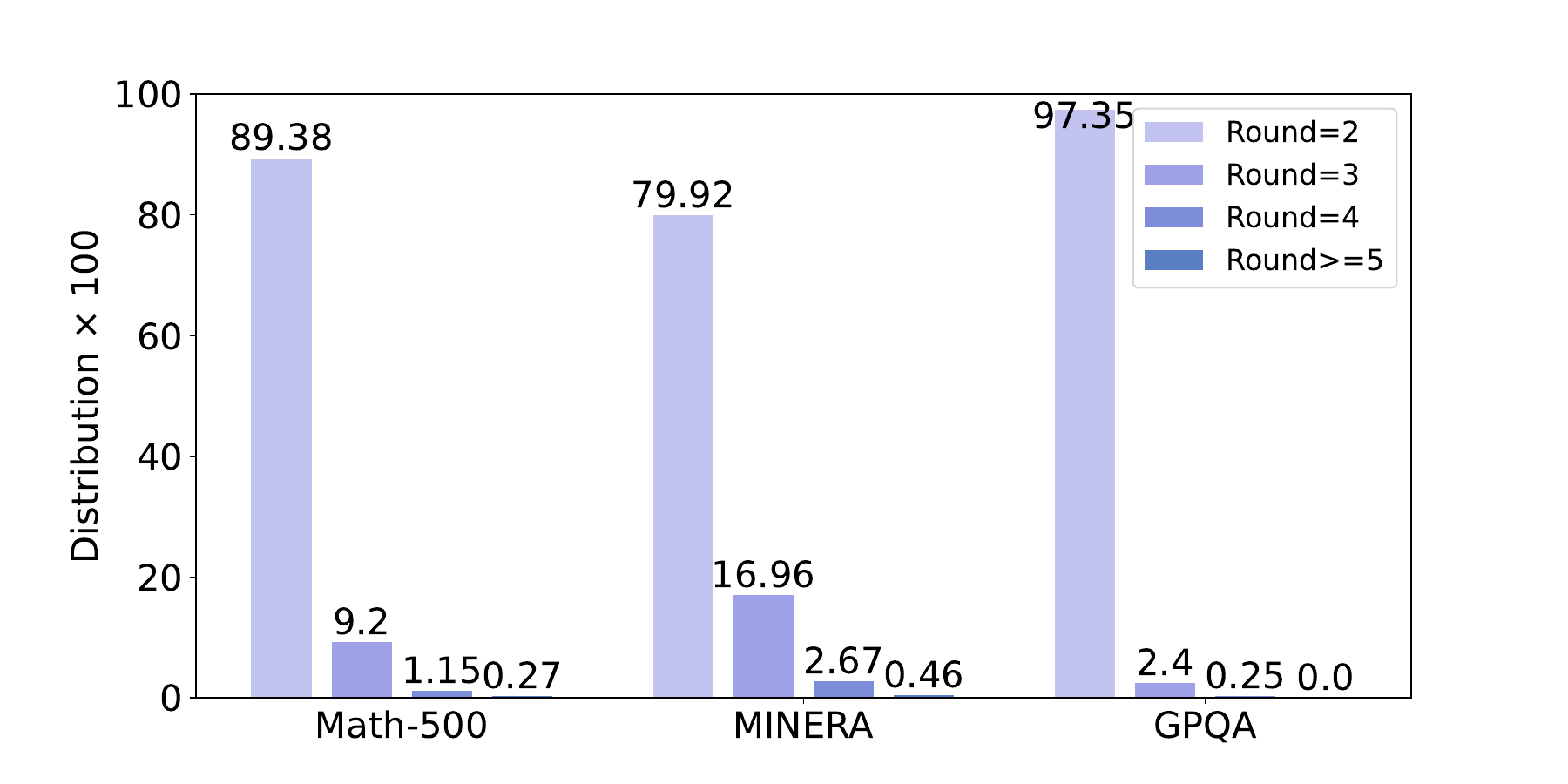}
			%\caption{fig1}
		\end{minipage}%
	}%
	\caption{Round distribution across models and datasets during different parallel settings with adaptive termination.}
        \label{fig:RD}
\end{figure}

\subsection{Analysis of the Number of Inference Round}
In this section, we analyze the inference round at which the model terminates generation. Fig.\ref{fig:RD} shows the distribution of stopping rounds for varied model scales, parallel configurations, and datasets. Please note that termination occurs no earlier than round 2, as round 1 is reserved for evaluating the dynamic SE baseline. It can be observed from Fig.\ref{fig:RD} that over 70\% of inferences terminate at the second round. This reveals that the model can successfully leverage the output from the first round to refine new responses and reduce uncertainty, as evidenced by the lower SE scores in round 2 versus round 1. Furthermore, our proposed adaptive termination mechanism finishes inference within 3 rounds in most scenarios, thereby avoiding the heavy computational cost ($O(N×M)$) of running all sequential steps. Moreover, scaling model size or increasing parallelization degree (higher $N$) will amplify second-round termination rates. This scaling behavior confirms that greater model capacity or expanded parallel search reinforces self-refinement capabilities, yielding higher-quality responses through earlier convergence.

\begin{figure}[!t]  % 允许 here / top / bottom / 新页    
\centering
    \includegraphics[width=0.75\linewidth]{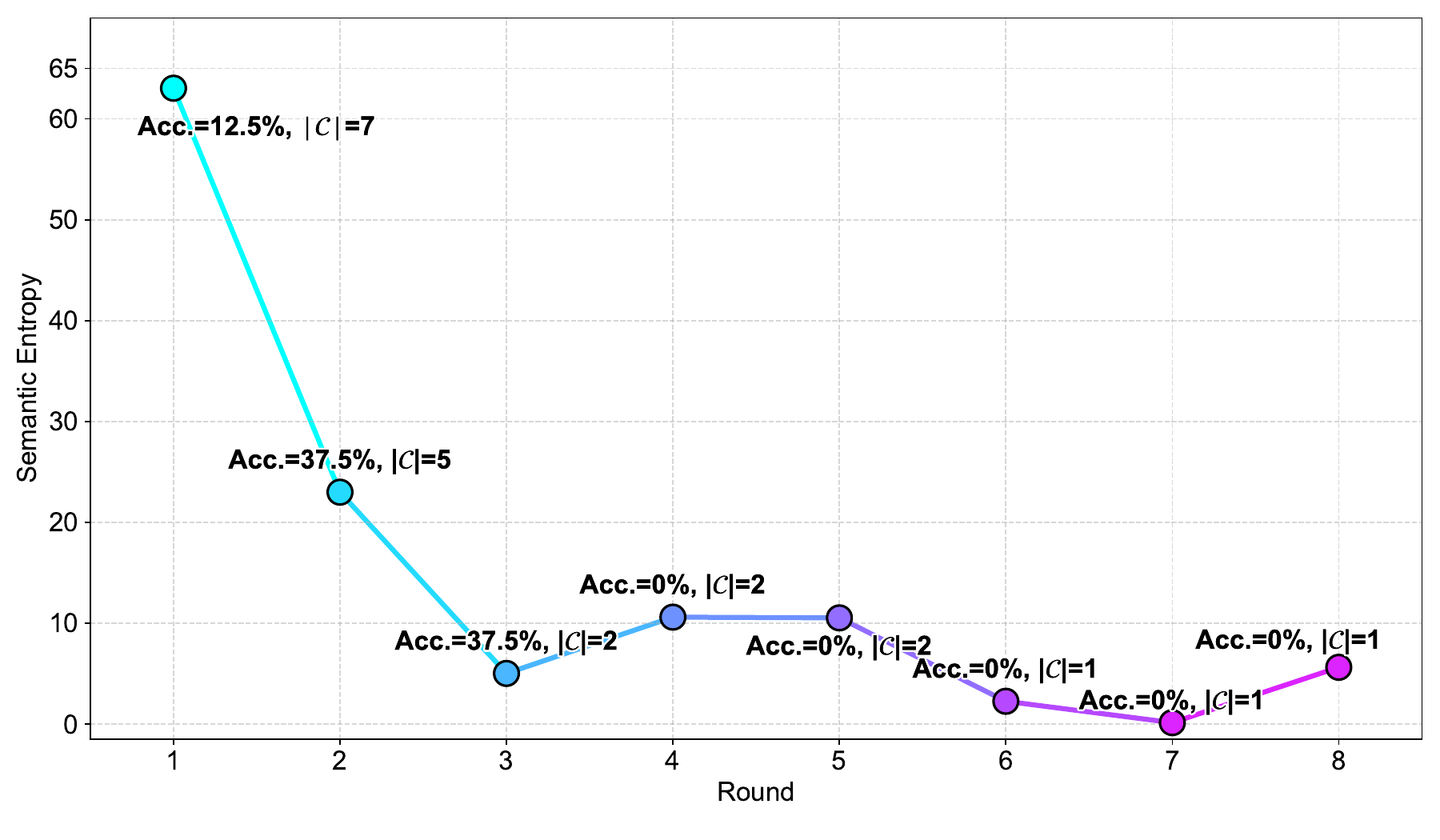} % 
    \caption{The evolution of SE and expect accuracy across different inference rounds of R7B
     model.}
\label{fig:collapse}
\end{figure}

\subsection{Visualization of SE Collapse} \label{sec:VSC}
This section will investigate an observation encountered during our experiments where, in the R7B model under $N=8$ configuration, the minimum selection yielded lower accuracy than the baseline. To investigate this phenomenon, we analyzed a representative case, tracking the evolution of SE and Expect accuracy, i.e., $\mathbb{E}(Acc.)$ across different inference rounds. As depicted in Fig.~\ref{fig:collapse}, while SE exhibits a pronounced decreasing trend, declining significantly from 63.04 at round-1 to a minimum of 0.14 by round-7, $\mathbb{E}(Acc.)$ first increases from 12.5 to 37.5 before plummeting to 0 at the later round. Meanwhile, the number of semantic clusters (i.e., $|\mathcal{C}|$) tends to decline throughout the inference process. This reduction directly reflects a loss of reasoning diversity, leading the model to respond overconfidently even when the answer is wrong. Specifically, upon examining representative outputs at round-7, we notice a remarkable reduction in response length. The model often skips detailed reasoning steps, directly outputting concise answers (specific examples are provided in the Appendix~\ref{app:eofsec}). We define this degradation as semantic entropy collapse, where the model exhibits precipitous SE deterioration in reasoning rounds, resulting in vanishing diversity and blindly outputting. We believe this phenomenon likely stems from the limited reasoning power of the model on a smaller scale (e.g. 7B), as performance for minimum SE rarely fell below baseline in the R32B model. Surprisingly, in this case, our proposed adaptive termination strategy stopped at round-2, achieving the highest expected accuracy, avoiding SE collapse, and effectively maintaining the small model's performance.

\section{Related Work}
Test-time scaling allows large language models (LLMs) to engage in more deliberative reasoning before producing final answers. Existing approaches can be broadly categorized into three main strategies, as discussed below.

\paragraph{Parallel Scaling.} 
Parallel scaling enables LLMs to independently generate multiple outputs for a given prompt, with the computational budget determined by the number of samples. To determine the final output, one line of research adopts unsupervised selection methods, typically relying on majority voting, to identify the most consistent output among the generated samples. This approach has demonstrated effectiveness across various reasoning tasks \citep{Wang2022SelfConsistencyIC, Chen2023UniversalSF}. However, its performance tends to saturate beyond a few hundred samples, showing limited improvement with additional sampling \citep{brown2024large}. Another line of work incorporates external verifiers to evaluate and select from the sampled outputs, achieving promising results in mathematical problem solving \citep{Cobbe2021TrainingVT, uesato2022solving, lightman2024lets} and program synthesis \citep{brown2024large}. Although parallel scaling facilitates diverse exploration by producing independent samples, it lacks coordination across different sampling trajectories, which limits its ability to support iterative refinement.

\paragraph{Sequential Scaling.}
Sequential scaling enables LLMs to engage in a long chain-of-thought reasoning process before producing a final answer, incorporating cognitive behaviors such as verification, backtracking, and subgoal decomposition \citep{gandhi2025cognitive}. This approach has driven notable progress in complex reasoning tasks, as evidenced by state-of-the-art models such as OpenAI O1 \citep{jaech2024openai} and DeepSeek R1 \citep{DeepSeekAI2025DeepSeekR1IR}. To regulate the computational budget of sequential scaling, researchers often introduce control signals through special tokens (e.g., ``wait'' or ``Final answer'') within the reasoning process \citep{Muennighoff2025s1ST, AlphaOne25}. Beyond single-pass generation, \citet{Tian2025ThinkTE} propose a multi-round approach, in which answers from earlier rounds are fed back into the model for further reasoning to enable iterative refinement. By leveraging informative intermediate steps, sequential scaling encourages models to perform more deliberate and focused reasoning. However, LLMs frequently become trapped in incorrect reasoning trajectories and face challenges in recovering to reach correct answers \citep{Zeng2025RevisitingTT, Luo2025LearningFP}.

\paragraph{Hybrid Scaling.}
Recent research has explored hybrid scaling strategies that aim to integrate the complementary strengths of parallel scaling (diverse exploration) and sequential scaling (conditioning on intermediate information). \citet{Pan2025LearningAP} propose adaptive parallel reasoning, in which a parent thread decomposes the task into subgoals and dispatches them to multiple child threads. The child threads then return summaries of their reasoning processes, which the parent thread subsequently aggregates for further sequential reasoning. \citet{Luo2025LearningFP} initiate reasoning with multiple independent chain-of-thought processes and introduce a routing mechanism that enables information exchange among these processes. The model benefits from parallel exploration, subsequently performing self-verification and iterative refinement based on the shared information. Distinct from previous approaches that explicitly intervene in the generation process, this work adopts a less intrusive and more flexible paradigm. Reasoning begins with several independent parallel trajectories,  and the diverse intermediate results obtained in the first stage provide valuable guidance for refinement in subsequent rounds. Furthermore, an unsupervised indicator based on semantic entropy is introduced to adaptively determine the number of sequential reasoning steps, allowing the model to dynamically allocate computational resources based on task complexity.

\section{Conclusion}
\label{sec:conclusion}
This work proposes SEAT, a test-time scaling framework for large language models reasoning that leverages the semantic entropy of model responses to enable synergy between parallel exploration and sequential refinement. We demonstrate the critical role of semantic entropy in measuring the parallel reasoning performance of the model. Guided by the semantic entropy of multiple parallel responses in each round, SEAT dynamically adapts the depth of the sequential reasoning. Extensive evaluation across five challenging benchmarks corroborates that SEAT significantly enhances the LLMs' reasoning performance. Furthermore, SEAT's dynamic termination strategy effectively avoids semantic entropy collapses potentially incurred by compact LLMs during parallel scaling with sequential steps. 

% fine-grain

% verifier

\bibliography{iclr2025_conference}

\begin{thebibliography}{33}
\providecommand{\natexlab}[1]{#1}
\providecommand{\url}[1]{\texttt{#1}}
\expandafter\ifx\csname urlstyle\endcsname\relax
  \providecommand{\doi}[1]{doi: #1}\else
  \providecommand{\doi}{doi: \begingroup \urlstyle{rm}\Url}\fi

\bibitem[Brown et~al.(2024)Brown, Juravsky, Ehrlich, Clark, Le, R{\'e}, and Mirhoseini]{brown2024large}
Bradley Brown, Jordan Juravsky, Ryan Ehrlich, Ronald Clark, Quoc~V Le, Christopher R{\'e}, and Azalia Mirhoseini.
\newblock Large language monkeys: Scaling inference compute with repeated sampling.
\newblock \emph{arXiv preprint arXiv:2407.21787}, 2024.

\bibitem[Chen et~al.(2025)Chen, Chen, Wang, and Yang]{Chen2025SEEDGRPOSE}
Minghan Chen, Guikun Chen, Wenguan Wang, and Yi~Yang.
\newblock Seed-grpo: Semantic entropy enhanced grpo for uncertainty-aware policy optimization.
\newblock \emph{ArXiv}, abs/2505.12346, 2025.
\newblock URL \url{https://api.semanticscholar.org/CorpusID:278741064}.

\bibitem[Chen et~al.(2024)Chen, Xu, Liang, He, Pang, Yu, Song, Liu, Zhou, Zhang, et~al.]{chen2024not}
Xingyu Chen, Jiahao Xu, Tian Liang, Zhiwei He, Jianhui Pang, Dian Yu, Linfeng Song, Qiuzhi Liu, Mengfei Zhou, Zhuosheng Zhang, et~al.
\newblock Do not think that much for 2+ 3=? on the overthinking of o1-like llms.
\newblock \emph{arXiv preprint arXiv:2412.21187}, 2024.

\bibitem[Chen et~al.(2023)Chen, Aksitov, Alon, Ren, Xiao, Yin, Prakash, Sutton, Wang, and Zhou]{Chen2023UniversalSF}
Xinyun Chen, Renat Aksitov, Uri Alon, Jie Ren, Kefan Xiao, Pengcheng Yin, Sushant Prakash, Charles Sutton, Xuezhi Wang, and Denny Zhou.
\newblock Universal self-consistency for large language model generation.
\newblock \emph{ArXiv}, abs/2311.17311, 2023.
\newblock URL \url{https://api.semanticscholar.org/CorpusID:265498407}.

\bibitem[Cobbe et~al.(2021)Cobbe, Kosaraju, Bavarian, Chen, Jun, Kaiser, Plappert, Tworek, Hilton, Nakano, Hesse, and Schulman]{Cobbe2021TrainingVT}
Karl Cobbe, Vineet Kosaraju, Mo~Bavarian, Mark Chen, Heewoo Jun, Lukasz Kaiser, Matthias Plappert, Jerry Tworek, Jacob Hilton, Reiichiro Nakano, Christopher Hesse, and John Schulman.
\newblock Training verifiers to solve math word problems.
\newblock \emph{ArXiv}, abs/2110.14168, 2021.
\newblock URL \url{https://api.semanticscholar.org/CorpusID:239998651}.

\bibitem[DeepSeek-AI et~al.(2025)DeepSeek-AI, Guo, Yang, Zhang, Song, Zhang, Xu, Zhu, Ma, Wang, Bi, Zhang, Yu, Wu, Wu, Gou, Shao, Li, Gao, Liu, Xue, Wang, Wu, Feng, Lu, Zhao, Deng, Zhang, Ruan, Dai, Chen, Ji, Li, Lin, Dai, Luo, Hao, Chen, Li, Zhang, Bao, Xu, Wang, Ding, Xin, Gao, Qu, Li, Guo, Li, Wang, Chen, Yuan, Qiu, Li, Cai, Ni, Liang, Chen, Dong, Hu, Gao, Guan, Huang, Yu, Wang, Zhang, Zhao, Wang, Zhang, Xu, Xia, Zhang, Zhang, Tang, Li, Wang, Li, Tian, Huang, Zhang, Wang, Chen, Du, Ge, Zhang, Pan, Wang, Chen, Jin, Chen, Lu, Zhou, Chen, Ye, Wang, Yu, Zhou, Pan, Li, Zhou, Wu, Yun, Pei, Sun, Wang, Zeng, Zhao, Liu, Liang, Gao, Yu, Zhang, Xiao, An, Liu, Wang, aokang Chen, Nie, Cheng, Liu, Xie, Liu, Yang, Li, Su, Lin, Li, Jin, Shen, Chen, Sun, Wang, Song, Zhou, Wang, Shan, Li, Wang, Wei, Zhang, Xu, Li, Zhao, Sun, Wang, Yu, Zhang, Shi, Xiong, He, Piao, Wang, Tan, Ma, Liu, Guo, Ou, Wang, Gong, Zou, He, Xiong, Luo, mei You, Liu, Zhou, Zhu, Huang, Li, Zheng, Zhu, Ma, Tang, Zha, Yan, Ren, Ren, Sha, Fu, Xu, Xie, guo
  Zhang, Hao, Ma, Yan, Wu, Gu, Zhu, Liu, Li, Xie, Song, Pan, Huang, Xu, Zhang, and Zhang]{DeepSeekAI2025DeepSeekR1IR}
DeepSeek-AI, Daya Guo, Dejian Yang, Haowei Zhang, Jun-Mei Song, Ruoyu Zhang, Runxin Xu, Qihao Zhu, Shirong Ma, Peiyi Wang, Xiaoling Bi, Xiaokang Zhang, Xingkai Yu, Yu~Wu, Z.~F. Wu, Zhibin Gou, Zhihong Shao, Zhuoshu Li, Ziyi Gao, Aixin Liu, Bing Xue, Bing-Li Wang, Bochao Wu, Bei Feng, Chengda Lu, Chenggang Zhao, Chengqi Deng, Chenyu Zhang, Chong Ruan, Damai Dai, Deli Chen, Dong-Li Ji, Erhang Li, Fangyun Lin, Fucong Dai, Fuli Luo, Guangbo Hao, Guanting Chen, Guowei Li, H.~Zhang, Han Bao, Hanwei Xu, Haocheng Wang, Honghui Ding, Huajian Xin, Huazuo Gao, Hui Qu, Hui Li, Jianzhong Guo, Jiashi Li, Jiawei Wang, Jingchang Chen, Jingyang Yuan, Junjie Qiu, Junlong Li, Jiong Cai, Jiaqi Ni, Jian Liang, Jin Chen, Kai Dong, Kai Hu, Kaige Gao, Kang Guan, Kexin Huang, Kuai Yu, Lean Wang, Lecong Zhang, Liang Zhao, Litong Wang, Liyue Zhang, Lei Xu, Leyi Xia, Mingchuan Zhang, Minghua Zhang, M.~Tang, Meng Li, Miaojun Wang, Mingming Li, Ning Tian, Panpan Huang, Peng Zhang, Qiancheng Wang, Qinyu Chen, Qiushi Du, Ruiqi Ge, Ruisong
  Zhang, Ruizhe Pan, Runji Wang, R.~J. Chen, Ruiqi Jin, Ruyi Chen, Shanghao Lu, Shangyan Zhou, Shanhuang Chen, Shengfeng Ye, Shiyu Wang, Shuiping Yu, Shunfeng Zhou, Shuting Pan, S.~S. Li, Shuang Zhou, Shao-Kang Wu, Tao Yun, Tian Pei, Tianyu Sun, T.~Wang, Wangding Zeng, Wanjia Zhao, Wen Liu, Wenfeng Liang, Wenjun Gao, Wen-Xia Yu, Wentao Zhang, Wangding Xiao, Wei An, Xiaodong Liu, Xiaohan Wang, Xi~aokang Chen, Xiaotao Nie, Xin Cheng, Xin Liu, Xin Xie, Xingchao Liu, Xinyu Yang, Xinyuan Li, Xuecheng Su, Xuheng Lin, X.~Q. Li, Xiangyu Jin, Xi-Cheng Shen, Xiaosha Chen, Xiaowen Sun, Xiaoxiang Wang, Xinnan Song, Xinyi Zhou, Xianzu Wang, Xinxia Shan, Y.~K. Li, Y.~Q. Wang, Y.~X. Wei, Yang Zhang, Yanhong Xu, Yao Li, Yao Zhao, Yaofeng Sun, Yaohui Wang, Yi~Yu, Yichao Zhang, Yifan Shi, Yi~Xiong, Ying He, Yishi Piao, Yisong Wang, Yixuan Tan, Yiyang Ma, Yiyuan Liu, Yongqiang Guo, Yuan Ou, Yuduan Wang, Yue Gong, Yu-Jing Zou, Yujia He, Yunfan Xiong, Yu-Wei Luo, Yu~mei You, Yuxuan Liu, Yuyang Zhou, Y.~X. Zhu, Yanping Huang, Yao
  Li, Yi~Zheng, Yuchen Zhu, Yunxiang Ma, Ying Tang, Yukun Zha, Yuting Yan, Zehui Ren, Zehui Ren, Zhangli Sha, Zhe Fu, Zhean Xu, Zhenda Xie, Zhen guo Zhang, Zhewen Hao, Zhicheng Ma, Zhigang Yan, Zhiyu Wu, Zihui Gu, Zijia Zhu, Zijun Liu, Zi-An Li, Ziwei Xie, Ziyang Song, Zizheng Pan, Zhen Huang, Zhipeng Xu, Zhongyu Zhang, and Zhen Zhang.
\newblock Deepseek-r1: Incentivizing reasoning capability in llms via reinforcement learning.
\newblock \emph{ArXiv}, abs/2501.12948, 2025.
\newblock URL \url{https://api.semanticscholar.org/CorpusID:275789950}.

\bibitem[Farquhar et~al.(2024)Farquhar, Kossen, Kuhn, and Gal]{Farquhar2024DetectingHI}
Sebastian Farquhar, Jannik Kossen, Lorenz Kuhn, and Yarin Gal.
\newblock Detecting hallucinations in large language models using semantic entropy.
\newblock \emph{Nature}, 630:\penalty0 625 -- 630, 2024.
\newblock URL \url{https://api.semanticscholar.org/CorpusID:270615909}.

\bibitem[Ferguson(1989)]{Ferguson1989WhoST}
Thomas~S. Ferguson.
\newblock Who solved the secretary problem.
\newblock \emph{Statistical Science}, 4:\penalty0 282--289, 1989.
\newblock URL \url{https://api.semanticscholar.org/CorpusID:62172567}.

\bibitem[Gandhi et~al.(2025)Gandhi, Chakravarthy, Singh, Lile, and Goodman]{gandhi2025cognitive}
Kanishk Gandhi, Ayush Chakravarthy, Anikait Singh, Nathan Lile, and Noah~D Goodman.
\newblock Cognitive behaviors that enable self-improving reasoners, or, four habits of highly effective stars.
\newblock \emph{arXiv preprint arXiv:2503.01307}, 2025.

\bibitem[Hendrycks et~al.(2021)Hendrycks, Burns, Kadavath, Arora, Basart, Tang, Song, and Steinhardt]{Hendrycks2021MeasuringMP}
Dan Hendrycks, Collin Burns, Saurav Kadavath, Akul Arora, Steven Basart, Eric Tang, Dawn~Xiaodong Song, and Jacob Steinhardt.
\newblock Measuring mathematical problem solving with the math dataset.
\newblock \emph{NeurIPS Datasets and Benchmarks}, 2021.

\bibitem[Jaech et~al.(2024)Jaech, Kalai, Lerer, Richardson, El-Kishky, Low, Helyar, Madry, Beutel, Carney, et~al.]{jaech2024openai}
Aaron Jaech, Adam Kalai, Adam Lerer, Adam Richardson, Ahmed El-Kishky, Aiden Low, Alec Helyar, Aleksander Madry, Alex Beutel, Alex Carney, et~al.
\newblock Openai o1 system card.
\newblock \emph{arXiv preprint arXiv:2412.16720}, 2024.

\bibitem[Kang et~al.(2025)Kang, Zhao, and Song]{Kang2025ScalableBS}
Zhewei Kang, Xuandong Zhao, and Dawn~Xiaodong Song.
\newblock Scalable best-of-n selection for large language models via self-certainty.
\newblock \emph{ArXiv}, abs/2502.18581, 2025.
\newblock URL \url{https://api.semanticscholar.org/CorpusID:276618155}.

\bibitem[Kuhn et~al.(2023)Kuhn, Gal, and Farquhar]{Kuhn2023SemanticUL}
Lorenz Kuhn, Yarin Gal, and Sebastian Farquhar.
\newblock Semantic uncertainty: Linguistic invariances for uncertainty estimation in natural language generation.
\newblock \emph{ArXiv}, abs/2302.09664, 2023.
\newblock URL \url{https://api.semanticscholar.org/CorpusID:257039062}.

\bibitem[Lewkowycz et~al.(2022)Lewkowycz, Andreassen, Dohan, Dyer, Michalewski, Ramasesh, Slone, Anil, Schlag, Gutman-Solo, et~al.]{lewkowycz2022solving}
Aitor Lewkowycz, Anders Andreassen, David Dohan, Ethan Dyer, Henryk Michalewski, Vinay Ramasesh, Ambrose Slone, Cem Anil, Imanol Schlag, Theo Gutman-Solo, et~al.
\newblock Solving quantitative reasoning problems with language models.
\newblock \emph{Advances in Neural Information Processing Systems}, 35:\penalty0 3843--3857, 2022.

\bibitem[Liang et~al.(2024)Liang, Song, Zheng, Wang, Yu, Li, Li, Xiong, and Li]{Liang2024InternalCA}
Xun Liang, Shichao Song, Zifan Zheng, Hanyu Wang, Qingchen Yu, Xunkai Li, Rong-Hua Li, Feiyu Xiong, and Zhiyu Li.
\newblock Internal consistency and self-feedback in large language models: A survey.
\newblock \emph{ArXiv}, abs/2407.14507, 2024.
\newblock URL \url{https://api.semanticscholar.org/CorpusID:271310469}.

\bibitem[Lightman et~al.(2024)Lightman, Kosaraju, Burda, Edwards, Baker, Lee, Leike, Schulman, Sutskever, and Cobbe]{lightman2024lets}
Hunter Lightman, Vineet Kosaraju, Yuri Burda, Harrison Edwards, Bowen Baker, Teddy Lee, Jan Leike, John Schulman, Ilya Sutskever, and Karl Cobbe.
\newblock Let's verify step by step.
\newblock In \emph{The Twelfth International Conference on Learning Representations}, 2024.

\bibitem[Luo et~al.(2025)Luo, Du, Bi, Chung, Tang, Yang, Zhang, and Wang]{Luo2025LearningFP}
Tongxu Luo, Wenyu Du, Jiaxi Bi, Stephen Chung, Zhengyang Tang, Hao Yang, Min Zhang, and Benyou Wang.
\newblock Learning from peers in reasoning models.
\newblock \emph{ArXiv}, abs/2505.07787, 2025.
\newblock URL \url{https://api.semanticscholar.org/CorpusID:278534473}.

\bibitem[Malinin \& Gales(2021)Malinin and Gales]{Malinin2021UncertaintyEI}
Andrey Malinin and Mark John~Francis Gales.
\newblock Uncertainty estimation in autoregressive structured prediction.
\newblock In \emph{International Conference on Learning Representations}, 2021.
\newblock URL \url{https://api.semanticscholar.org/CorpusID:231895728}.

\bibitem[Muennighoff et~al.(2025)Muennighoff, Yang, Shi, Li, Li, Hajishirzi, Zettlemoyer, Liang, Candes, and Hashimoto]{Muennighoff2025s1ST}
Niklas Muennighoff, Zitong Yang, Weijia Shi, Xiang~Lisa Li, Fei-Fei Li, Hanna Hajishirzi, Luke~S. Zettlemoyer, Percy Liang, Emmanuel~J. Candes, and Tatsunori Hashimoto.
\newblock s1: Simple test-time scaling.
\newblock \emph{ArXiv}, abs/2501.19393, 2025.
\newblock URL \url{https://api.semanticscholar.org/CorpusID:276079693}.

\bibitem[OpenAI(2024)]{o1}
OpenAI.
\newblock Learning to reason with llms.
\newblock \url{https://openai.com/index/learning-to-reason-with-llms/}, 2024.

\bibitem[Pan et~al.(2025)Pan, Li, Lian, Snell, Zhou, Yala, Darrell, Keutzer, and Suhr]{Pan2025LearningAP}
Jiayi Pan, Xiuyu Li, Long Lian, Charlie Snell, Yifei Zhou, Adam Yala, Trevor Darrell, Kurt Keutzer, and Alane Suhr.
\newblock Learning adaptive parallel reasoning with language models.
\newblock \emph{ArXiv}, abs/2504.15466, 2025.
\newblock URL \url{https://api.semanticscholar.org/CorpusID:277994172}.

\bibitem[Rein et~al.(2023)Rein, Hou, Stickland, Petty, Pang, Dirani, Michael, and Bowman]{Rein2023GPQAAG}
David Rein, Betty~Li Hou, Asa~Cooper Stickland, Jackson Petty, Richard~Yuanzhe Pang, Julien Dirani, Julian Michael, and Samuel~R. Bowman.
\newblock Gpqa: A graduate-level google-proof q\&a benchmark.
\newblock \emph{ArXiv}, abs/2311.12022, 2023.
\newblock URL \url{https://api.semanticscholar.org/CorpusID:265295009}.

\bibitem[Shiryaev(1980)]{Shiryaev1980OptimalSR}
Albert~N. Shiryaev.
\newblock Optimal stopping rules.
\newblock In \emph{International Encyclopedia of Statistical Science}, 1980.
\newblock URL \url{https://api.semanticscholar.org/CorpusID:46379699}.

\bibitem[Snell et~al.(2025)Snell, Lee, Xu, and Kumar]{Snell2025ScalingLT}
Charlie~Victor Snell, Jaehoon Lee, Kelvin Xu, and Aviral Kumar.
\newblock Scaling llm test-time compute optimally can be more effective than scaling parameters for reasoning.
\newblock In \emph{International Conference on Learning Representations}, 2025.
\newblock URL \url{https://api.semanticscholar.org/CorpusID:278498044}.

\bibitem[Team(2025)]{qwq}
Qwen Team.
\newblock Qwq-32b: Embracing the power of reinforcement learning.
\newblock \url{https://qwenlm.github.io/blog/qwq-32b/}, 2025.

\bibitem[Tian et~al.(2025)Tian, Zhao, Wang, Chen, Ji, Peng, Zhao, and Li]{Tian2025ThinkTE}
Xiaoyu Tian, Sitong Zhao, Haotian Wang, Shuaiting Chen, Yunjie Ji, Yiping Peng, Han Zhao, and Xiangang Li.
\newblock Think twice: Enhancing llm reasoning by scaling multi-round test-time thinking.
\newblock \emph{ArXiv}, abs/2503.19855, 2025.
\newblock URL \url{https://api.semanticscholar.org/CorpusID:277314100}.

\bibitem[Uesato et~al.(2022)Uesato, Kushman, Kumar, Song, Siegel, Wang, Creswell, Irving, and Higgins]{uesato2022solving}
Jonathan Uesato, Nate Kushman, Ramana Kumar, Francis Song, Noah Siegel, Lisa Wang, Antonia Creswell, Geoffrey Irving, and Irina Higgins.
\newblock Solving math word problems with process-and outcome-based feedback.
\newblock \emph{arXiv preprint arXiv:2211.14275}, 2022.

\bibitem[Wang et~al.(2025{\natexlab{a}})Wang, Yu, Gao, Zheng, Liu, Lu, Dang, Chen, Yang, Zhang, Liu, Yang, Zhao, Yue, Song, Yu, Huang, and Lin]{Wang2025BeyondT8}
Shenzhi Wang, Le~Yu, Chang Gao, Chujie Zheng, Shixuan Liu, Rui Lu, Kai Dang, Xionghui Chen, Jianxin Yang, Zhenru Zhang, Yuqiong Liu, An~Yang, Andrew Zhao, Yang Yue, Shiji Song, Bowen Yu, Gao Huang, and Junyang Lin.
\newblock Beyond the 80/20 rule: High-entropy minority tokens drive effective reinforcement learning for llm reasoning.
\newblock 2025{\natexlab{a}}.
\newblock URL \url{https://api.semanticscholar.org/CorpusID:279119146}.

\bibitem[Wang et~al.(2022)Wang, Wei, Schuurmans, Le, Chi, and Zhou]{Wang2022SelfConsistencyIC}
Xuezhi Wang, Jason Wei, Dale Schuurmans, Quoc Le, Ed~H. Chi, and Denny Zhou.
\newblock Self-consistency improves chain of thought reasoning in language models.
\newblock \emph{International Conference on Learning Representations}, 2022.

\bibitem[Wang et~al.(2025{\natexlab{b}})Wang, Liu, Xu, Liang, Chen, He, Song, Yu, Li, Zhang, et~al.]{wang2025thoughts}
Yue Wang, Qiuzhi Liu, Jiahao Xu, Tian Liang, Xingyu Chen, Zhiwei He, Linfeng Song, Dian Yu, Juntao Li, Zhuosheng Zhang, et~al.
\newblock Thoughts are all over the place: On the underthinking of o1-like llms.
\newblock \emph{arXiv preprint arXiv:2501.18585}, 2025{\natexlab{b}}.

\bibitem[Welleck et~al.(2024)Welleck, Bertsch, Finlayson, Schoelkopf, Xie, Neubig, Kulikov, and Harchaoui]{Welleck2024FromDT}
Sean Welleck, Amanda Bertsch, Matthew Finlayson, Hailey Schoelkopf, Alex Xie, Graham Neubig, Ilia Kulikov, and Zaid Harchaoui.
\newblock From decoding to meta-generation: Inference-time algorithms for large language models.
\newblock \emph{ArXiv}, abs/2406.16838, 2024.
\newblock URL \url{https://api.semanticscholar.org/CorpusID:270703266}.

\bibitem[Zeng et~al.(2025)Zeng, Cheng, Yin, Zhou, and Qiu]{Zeng2025RevisitingTT}
Zhiyuan Zeng, Qinyuan Cheng, Zhangyue Yin, Yunhua Zhou, and Xipeng Qiu.
\newblock Revisiting the test-time scaling of o1-like models: Do they truly possess test-time scaling capabilities?
\newblock \emph{ArXiv}, abs/2502.12215, 2025.
\newblock URL \url{https://api.semanticscholar.org/CorpusID:276421903}.

\bibitem[Zhang et~al.(2025)Zhang, Dong, Wang, Ning, Geng, Li, He, Bai, Malik, Gupta, and Zhang]{AlphaOne25}
Junyu Zhang, Runpei Dong, Han Wang, Xuying Ning, Haoran Geng, Peihao Li, Xialin He, Yutong Bai, Jitendra Malik, Saurabh Gupta, and Huan Zhang.
\newblock Alphaone: Reasoning models thinking slow and fast at test time.
\newblock \emph{arXiv preprint arXiv:2505.24863}, 2025.

\end{thebibliography}
\bibliographystyle{iclr2025_conference}

\appendix
\section{Example Outputs for Semantic Entropy Collapse} \label{app:eofsec}
The phenomenon of semantic entropy collapse can be seen through the following three illustrative examples.

\begin{AIbox}{Example 1}
...\textless\textbackslash think\textgreater

The inradius of triangle ABC is 35, expressed as the fraction 35/1. Thus, m + n equals 35 + 1, which is 36.

[ \\
\textbackslash boxed{36} \\
]
\end{AIbox}

\begin{AIbox}{Example 2}
...\textless\textbackslash think\textgreater

The inradius of triangle \(ABC\) is the sum of the radii of the two tangent circles, which are 34 and 1. Therefore, the inradius \(r = 34 + 1 = 35\). Expressed as a fraction, this is \(\frac{35}{1}\). The sum of the numerator and denominator is \(35 + 1 = 36\).

[ \\
\textbackslash boxed{36} \\
]
\end{AIbox}

\begin{AIbox}{Example 3}
...\textless\textbackslash think\textgreater

The inradius of triangle \(ABC\) is the sum of the radii of the two tangent circles, which are 34 and 1. Therefore, the inradius \(r = 34 + 1 = 35\). Expressed as a fraction, this is \(\frac{35}{1}\), so \(m = 35\) and \(n = 1\). Thus, \(m + n = 35 + 1 = 36\).

[ \\
\textbackslash boxed{36} \\
]
\end{AIbox}

\end{document}